\documentclass{article}

\usepackage{microtype}
\usepackage{graphicx}
\usepackage{subfigure}
\usepackage{booktabs} 
\usepackage{float}
\usepackage{amsmath}
\usepackage{amsfonts}
\usepackage{amsthm}
\usepackage{algorithmic}
\usepackage[ruled]{algorithm2e}
\usepackage{color}
\usepackage{multirow,booktabs}
\usepackage{xtab}
\usepackage{caption}
\usepackage{graphicx}
\usepackage{subfigure}
\usepackage{rotating}
\usepackage{lineno,hyperref}

\renewcommand{\algorithmicrequire}{\textbf{Input:}}
\renewcommand{\algorithmicensure}{\textbf{Output:}}

\newtheorem{proposition}{Proposition}[section]

\usepackage{hyperref}

\definecolor{Tianlong_color}{rgb}{0.858, 0.188, 0.478}

\usepackage[accepted]{icml2021}

\icmltitlerunning{Sparse and Imperceptible Adversarial Attack via a Homotopy Algorithm}

\begin{document}

\twocolumn[
\icmltitle{Sparse and Imperceptible Adversarial Attack via a Homotopy Algorithm}




\begin{icmlauthorlist}
\icmlauthor{Mingkang Zhu}{to}
\icmlauthor{Tianlong Chen}{to}
\icmlauthor{Zhangyang Wang}{to}

\end{icmlauthorlist}

\icmlaffiliation{to}{The University of Texas at Austin, USA}

\icmlcorrespondingauthor{Zhangyang Wang}{atlaswang@utexas.edu}

\icmlkeywords{Machine Learning, ICML}

\vskip 0.3in
]



\printAffiliationsAndNotice{} 

\begin{abstract}

Sparse adversarial attacks can fool deep neural networks (DNNs) by only perturbing a few pixels (regularized by $\ell_0$ norm). Recent efforts combine it with another $\ell_\infty$ imperceptible on the perturbation magnitudes. The resultant sparse and imperceptible attacks are practically relevant, and indicate an even higher vulnerability of DNNs that we usually imagined. However, such attacks are more challenging to generate due to the optimization difficulty by coupling the $\ell_0$ regularizer and box constraints with a non-convex objective. In this paper, we address this challenge by proposing a homotopy algorithm, to jointly tackle the sparsity and the perturbation bound in one unified framework. Each iteration, the main step of our algorithm is to optimize an $\ell_0$-regularized adversarial loss, by leveraging the nonmonotone Accelerated Proximal Gradient Method (nmAPG) for nonconvex programming; it is followed by an $\ell_0$ change control step, and an optional post-attack step designed to escape bad local minima. We also extend the algorithm to handling the structural sparsity regularizer. We extensively examine the effectiveness of our proposed \textbf{homotopy attack} for both targeted and non-targeted attack scenarios, on CIFAR-10 and ImageNet datasets. Compared to state-of-the-art methods, our homotopy attack leads to significantly fewer  perturbations, e.g., reducing  42.91\% on CIFAR-10 and 75.03\% on ImageNet (average case, targeted attack), at similar maximal perturbation magnitudes, when still achieving 100\% attack success rates. Our codes are available at: {\small\url{https://github.com/VITA-Group/SparseADV_Homotopy}}.
\end{abstract}

\section{Introduction}\label{submission}

Deep neural networks (DNNs) have widely demonstrated fragility to adversarial attacks, e.g., maliciously crafted small perturbations to their inputs that can fool them to make incorrect and implausible predictions \cite{DBLP:journals/corr/CarliniW16a, ijcai2018-543, pmlr-v80-athalye18b,zhang2020geometry,Zhou_2020_CVPR,gao2020maximum,zhang2020attacks,du2021learning}. Adversarial attacks raise serious concern against DNNs' applicability in high-stake, risk-sensitive applications. Meanwhile, probing and leveraging adversarial attacks could help us troubleshoot the weakness of a DNN, and further leading to strengthening it, e.g., by adversarial training \cite{madry2017towards}.

\begin{figure}[t]
\centering
     \subfigure [otterhound]
     {
        \includegraphics[height=1.8cm]{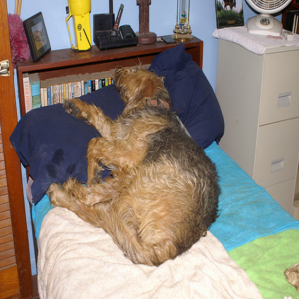}
    }
        \subfigure [stingray]
    {
        \includegraphics[height=1.8cm]{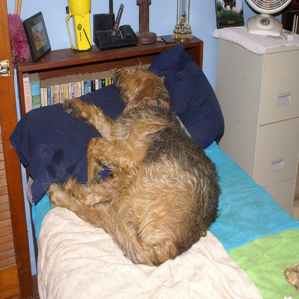}
    }
    \subfigure [$\ell_0$ = 1803]
    {
        \includegraphics[height=1.8cm]{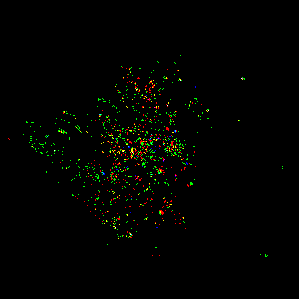}
    }
    \subfigure [$\ell_0$ = 12264]
    {
        \includegraphics[height=1.8cm]{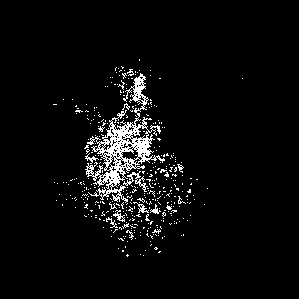}
    }

\vskip -0.3cm
     \subfigure [{\scriptsize American coot}]
     {
        \includegraphics[height=1.8cm]{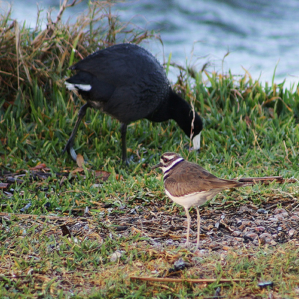}
    }
        \subfigure [stingray]
    {
        \includegraphics[height=1.8cm]{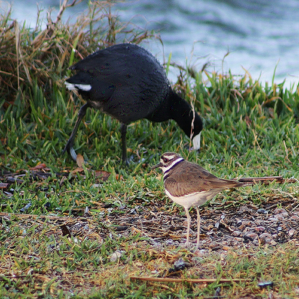}
    }
    \subfigure [$\ell_0$ = 1400]
    {
        \includegraphics[height=1.8cm]{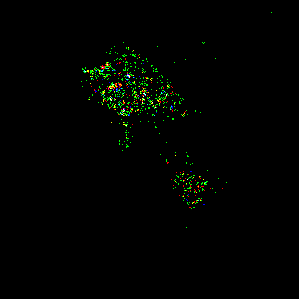}
    }
    \subfigure [$\ell_0$ = 13467]
    {
        \includegraphics[height=1.8cm]{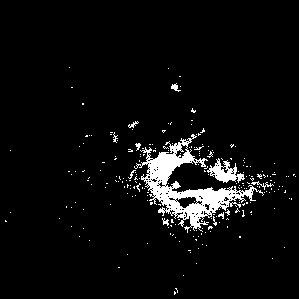}
    }

\vskip -0.3cm
     \subfigure [Boston bull]
     {
        \includegraphics[height=1.8cm]{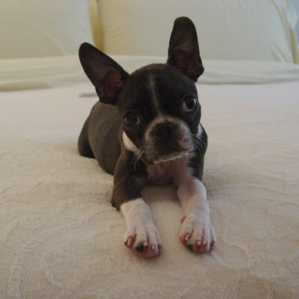}
    }
       \subfigure [stingray]
    {
        \includegraphics[height=1.8cm]{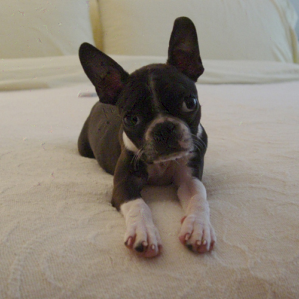}
    }
    \subfigure [$\ell_0$ = 2986]
    {
        \includegraphics[height=1.8cm]{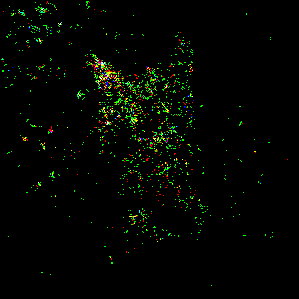}
    }
    \subfigure [$\ell_0$ = 16974]
    {
        \includegraphics[height=1.8cm]{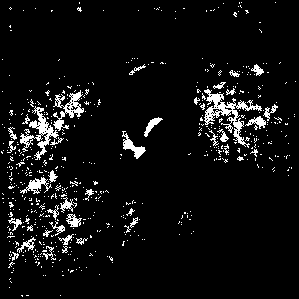}
    }

\vskip -0.3cm
  \caption{\small
      Visualization of pixel-wise sparsity of targeted attack when enforcing a $\ell_\infty$ constraint of 0.05. The images in each row, from the left to right are benign image, our adversarial example, our perturbation position, GreedyFool (a recent state-of-the-art approach in \citet{dong2020greedyfool})'s perturbation position. In the figures of perturbation positions, black pixels denote no perturbation; white pixels denote perturbing all three RGB channels; pure red, green, blue pixels represent perturbing single channel; and pixels with other color denote perturbing two channels.
  }\label{fig:pix}
\vspace{-4mm}
\end{figure}

A non-trivial adversarial attack is meant to be small and ``imperceptible" in certain sense \cite{nguyen2015deep}: in this way, it impacts little on the benign input's semantic meaning, hence uncovering the inherent instability of the DNN. In the most common pixel-level additive attack setting, the standard measure of attack magnitude is by the $\ell_p$-norm  of perturbations, with $p$ = 0, 1, 2, or {$\infty$}. In particular, a number of works \cite{Modas_2019_CVPR,Fan2020ECCV,xu2018structured,dong2020greedyfool,Croce_2019_ICCV} demonstrated that an attack can succeed by only perturbing a few pixels under the $\ell_0$ constraint (sometimes $\ell_1$). Fig.~\ref{fig:pix} shows examples of successful pixel-wise sparse targeted attack by our algorithm, compared with a recent state-of-the-art method GreedyFool \cite{dong2020greedyfool}. Further, since the vanilla $\ell_0$-attacks leave it unconstrained how they change each pixel, the perturbed pixels may have very different intensity or color than the surrounding ones and hence be easily visible \cite{DBLP:journals/corr/abs-1710-08864}. Therefore, more recent methods focus on enforcing some element-wise magnitude constraints to also ensure imperceptible perturbations. In total, such sparse and imperceptible adversarial attacks become a practically relevant topic of interest, indicating an even higher vulnerability of DNNs that we usually imagined, and providing additional insights about interpreting DNN failures \cite{xu2018structured}.



\subsection{Related Work}

Many works on generating sparse and imperceptible adversarial attacks have been proposed recently. The authors of \cite{Croce_2019_ICCV} proposed PGD$_0$ to project the PGD adversarial attack \cite{madry2017towards} to the $\ell_0$ ball, as well as a heuristic CornerSearch strategy by traversing all the pixels and to select a subset to  perturb.
SparseFool \cite{Modas_2019_CVPR} relaxes and approximates the 
$\ell_0$ objective using $\ell_1$ objective.  SAPF \cite{Fan2020ECCV} reformulates the sparse adversarial attack problem into a mixed integer
programming (MIP) problem, to jointly optimize the binary selection factors and continuous perturbation magnitudes of all pixels, with a cardinality constraint to explicitly control the degree of sparsity. It was solved by the $\ell_p$-Box ADMM designed for integer programming. StrAttack \cite{xu2018structured} focuses on group-wise sparsity and also solves the problem using an ADMM based algorithm. The latest algorithm, GreedyFool \cite{dong2020greedyfool}  presents a two-stage greedy solution that first picks perturbations with large gradients and then reduces useless perturbations in the second stage. Besides those white-box attacks, similar problems were also studied in the black-box setting, e.g., One Pixel \cite{DBLP:journals/corr/abs-1710-08864} and Pointwise Attack \cite{schott2018towards}.

Despite progress, challenges remain for those existing solutions. PGD$_0$ and CornerSearch may fail to yield full success attack rates \cite{Croce_2019_ICCV}; they moreover need pre-specified numbers of perturbed pixels, which can also cause inflexible and unnecessarily redundant perturbations. SparseFool \cite{Modas_2019_CVPR} did not directly control the number of perturbed pixels, and was designed for the nontargeted setting only. Among the latter optimization-based solutions, StrAttack \cite{xu2018structured} and SAPF \cite{Fan2020ECCV} suffer from slow and sometimes unstable convergence in practice, especially when the enforced sparsity is high; it is also not uncommon to observe GreedyFool \cite{dong2020greedyfool} be stuck in local minima because of its greedy nature.


Looking for a more principled optimization solution, we look back into the classical sparse optimization field intensively studied for decades. Among numerous methods proposed, nonconvex regularization based methods, including $\ell_0$ and  $\ell_p$-norm ($0<p<1$)-based, have achieved considerable performance improvement over the more traditional convex relaxation \cite{8531588}.  Specially, iterative hard thresholding methods or proximal gradient methods for $\ell_0$-regularized problems are shown to be scalable for (group) sparse regression \cite{NIPS2016_8e82ab72}. A thorough review of this field can be found in \cite{bach2011optimization}.

\subsection{Our Contributions}


This work establishes an $\ell_0$-regularized and $\ell_{\infty}$-constrained optimization framework, for finding sparse and imperceptible adversarial attacks. Our \textbf{technical foundations} are based on an innovative integration between the proximal gradient method for nonconvex optimization described in \cite{NIPS2015_f7664060}, and the homotopy algorithm.
In sparse convex optimization \cite{7084156, doi:10.1137/120869997}, the homotopy algorithm (a.k.a. continuation) keeps a sequence of decreasing weights for regularizers, yielding a multistage
homotopy solution. In each stage with a fixed weight, the $l_0$ or $l_1$-regularized convex optimization problem is optimized. That provides a sparse initial solution for the next stage optimization with a smaller weight, progressively leading to the final sparse solution.

The \textbf{core merit} of our method is to enable a smooth relaxation of the sparsity constraint, which eventually leads to a flexible yet compactly sparse solution. We outline our specific contributions below:\vspace{-1em}
\begin{itemize}
    \item We first introduce the nonmonotone Accelerated Proximal Gradient Method (nmAPG) to jointly tackle the sparsity and the box constraint. The algorithm can also be extended to handling \textit{group-wise sparsity}. A main hurdle in this regularization-based method is yet how to achieve the desired sparsity: this is typically controlled by some ad-hoc chosen regularization coefficient.\vspace{-0.2em}
    \item We then integrate the homotopy algorithm into our framework, which allows us to optimize using a sequence of gradually decreasing weights for regularizers \cite{7084156, doi:10.1137/120869997}. It can continuously provide a sparse initial solution for the next stage optimization, which leads to a compact sparse solution by the end. To deal with the weight sensitivity of the $\ell_0$-regularizer in the homotopy algorithm, we design an $\ell_0$-change control step on the perturbation update at each iteration, and another optional step called post-attack perturbation to escape bad local minima. Both lead to stable sparse solutions without extensive weight searching for regularization-based methods.\vspace{-0.2em}
    \item Extensive experiments on the CIFAR-10 \cite{Krizhevsky2009LearningML} and  ImageNet \cite{5206848} endorse the superiority of our new \textbf{homotopy attack}. In both targeted and non-targeted attack scenarios, it outperforms state-of-the-art methods by significantly fewer perturbations,
        e.g., reducing  42.91\% on CIFAR-10 and 75.03\% on ImageNet (average case, targeted attack), at similar perturbation magnitudes, when still achieving 100\% attack success rates. Visualizing the attacks also gains us more interpretation of DNN fragility.
\end{itemize}


%
%

\section{Method}

\subsection{Problem Formulation}

Let $\mathcal{X}$ be the set of benign images. A trained DNN image classification model maps each image $x \in \mathbb{R}^{n}$ from $\mathcal{X}$ to its class label $y\in \mathcal{Y}=\{1, 2, \cdots, K\}$. Given a benign image $x_0\in \mathcal{X}$, a \textit{targeted} sparse and imperceptible adversarial attack aims for finding a perturbation $\delta\in \mathbb{R}^n$, so that the perturbed image $x_0+\delta$ is incorrectly classified to a target class $t$. In the meanwhile, the number of  perturbations 
should be as sparse as possible, and the perturbation magnitude should remain small. This problem can be cast as
    \begin{equation}\begin{array}{ll}\label{equ1}
      \displaystyle\min_{\delta} & f(x_0+\delta, t) + \lambda ||\delta||_0\\
         s.t. & ||\delta||_{\infty}\le \epsilon, \,\, 0\le x_0+\delta\le 1.
    \end{array}\end{equation}
Here $f()$ is the classification loss such as the cross entropy function, $\lambda$ is the regularization coefficient to control the $\ell_0$ sparsity of $\delta$, and $\epsilon$ is the maximal allowable perturbation magnitude. This above targeted attack formulation can be easily adapted to the \textit{nontargeted} attack by replacing the adversarial loss $f()$:
    \begin{equation}\begin{array}{ll}\label{equ1-non}
      \displaystyle\min_{\delta} & f(x_0+\delta, y_0) + \lambda ||\delta||_0\\
         s.t. & ||\delta||_{\infty}\le \epsilon, \,\, 0\le x_0+\delta\le 1.
    \end{array}\end{equation}
Since the algorithms for solving \eqref{equ1} and \eqref{equ1-non} will have most in common, we will focus our discussion on \eqref{equ1} hereinafter, while noting that all algorithmic steps can be readily utilized towards solving \eqref{equ1-non} unless otherwise specified.

\subsection{A Basic Solution using Accelerated Proximal Gradient for Nonconvex Optimization}

Since both constraints in  \eqref{equ1} are box constraints, they can be simplified to one box constraint as $\delta\in [l, u]$, where $l, u\in \mathbb{R}^n$, and $0\in [l, u]$. Let $I_{[l, u]}$ be the indicator function
such that $I_{[l, u]}(\delta)=0$ if $\delta\in [l, u]$, and $I_{[l, u]}(\delta)= +\infty$ otherwise. Problem \eqref{equ1} can be rewritten as
      \begin{equation}\label{equ2}
          \displaystyle\min_{\delta}  f(x_0+\delta, t) + \lambda ||\delta||_0+I_{[l, u]}(\delta).
      \end{equation}
The non-convexity and non-smoothness make solving \eqref{equ2} a nontrivial task. The classical accelerated proximal gradient (APG) methods \cite{Nesterov04} have been proven to be efficient for convex optimization problems, but it remains unclear whether the usual APG can converge to a critical point of nonconvex programming. In \cite{NIPS2015_f7664060}, the authors proposed two convergent accelerated proximal gradient methods (APG) for  nonconvex and nonsmooth programs, of which \eqref{equ2} is a special case. In this paper, we adopt their proposed \textit{nonmonotone APG with line search} (nmAPG) algorithm.


To be self-containing, we briefly describe the high-level idea, and refer the readers to \cite{NIPS2015_f7664060} for more details. APG first obtains an extrapolation solution using the current and the previous solutions, and then solve proximal mapping problems. When extending APG to nonconvex programming, the main challenge is to prevent bad extrapolation due to oscillatory landscapes \cite{beck2009fast}. The authors of \cite{NIPS2015_f7664060} introduced a monitor to prevent and correct bad extrapolations so that they satisfy the sufficient descent property. In practice, nmAPG searches for a step size using the backtracking line search initialized with the BB step-length \cite{barzilai1988two} since the Lipstchiz constant of gradients is not exactly known.


We next lay out how to solve \eqref{equ2} with nmAPG. Suppose that $f$ is a smooth nonconvex function, whose gradient has  Lipschitz constant $L$. Given $\delta^k$, it holds that
      \begin{equation*}
      \begin{array}{l}
      f(x_0+\delta, t) \le f(x_0+\delta^k, t)+\\
      ~~~~~~~ \nabla_{\delta} f(x_0+\delta^k, t)^T(\delta-\delta^k)+ \frac L2 ||\delta-\delta^k||^2,
     \end{array}
      \end{equation*}
where $|| \cdot ||$ is the $|| \cdot ||_2$ norm for simplicity.

Then we consider the problem
       \begin{equation*}\begin{array}{ll}
          \displaystyle\min_{\delta} &  f(x_0+\delta^k, t)+ \nabla_{\delta} f(x_0+\delta^k, t)^T(\delta-\delta^k)+ \\ & \frac L2 ||\delta
           -\delta^k||^2+ \lambda ||\delta||_0+I_{[l, u]}(\delta),
          \end{array}
      \end{equation*}
which is equivalent to
      \begin{equation}\begin{array}{ll}\label{equ3}
          \displaystyle\min_{\delta} &\frac L2 ||\delta-[\delta^k-\frac 1L \nabla_{\delta} f(x_0+\delta^k, t)]||^2 + \\
          &\lambda ||\delta||_0+I_{[l, u]}(\delta).
          \end{array}
      \end{equation}
Now letting $g(\delta)=\lambda ||\delta||_0+I_{[l, u]}(\delta)$, we denote the solution to \eqref{equ3} as
    \begin{equation}\begin{array}{l} \label{prox}
    \text{Prox}_{\frac 1L g}(\delta^k-\frac 1L \nabla_{\delta} f(x_0+\delta^k, t)) =\arg\!\displaystyle\min_{\delta}  \frac L2 ||\delta-   [\delta^k-\\
    \frac 1L \nabla_{\delta} f(x_0+\delta^k, t)]||^2 + \lambda ||\delta||_0+I_{[l, u]}(\delta).
    \end{array}\end{equation}
In fact, it can be obtained explicitly. Let
$$S_L(\delta)=\delta-\frac 1L \nabla_{\delta} f(x_0+\delta, t), \forall \delta\in [l, u], $$
and
$$\Pi_{[l, u]}(\delta)=\arg\!\min\{||y-\delta||: y\in [l, u]\}, \forall \delta\in \mathbb{R}^n.$$
It is easy to obtained that {\cite{82e11be51f1e41209d87809aeccbe338}}, the optimal solution of problem \eqref{equ3} is (for $i=1, 2,\cdots, n$)
\begin{equation*}
      \delta_i^{k+1}=\left\{ \begin{array}{cl}
         [\Pi_{[l, u]}(s_L(\delta^k))]_i, & \text{~~if~} [s_L(\delta^k)]^2_i-  [\Pi_{[l, u]}(  \\
        & s_L(\delta^k))- s_L(\delta^k)]^2_i > \frac{2\lambda}L;\\
          0, & \text{~~otherwise,}\\
      \end{array}\right.
\end{equation*}
Since $g(\delta)=\lambda ||\delta||_0+I_{[l, u]}(\delta)$ is a proper and lower semicontinuous function, the convergence of nmAPG to a critical point of problem \eqref{equ2} can be assured, under some mild conditions {\cite{NIPS2015_f7664060}}.

\subsection{Extension to Group-wise Sparsity}

Group-wise sparsity 
has been recently incorporated into adversarial attack by the recent work \cite{xu2018structured} to provide more structured and explainable perturbations. To extend our algorithm from the element-wise $\ell_0$ regularizer to the group-wise sparsity, we need to reformulate the problem and adapt the nmAPG solution.


Suppose the input image $x_0$ can be partitioned into $m$ groups $x^T=(x^T_{G_1}, \cdots, x^T_{G_m})$. Then the problem concerning group-wise sparsity and imperceptibility can be cast as
    \begin{equation}\begin{array}{ll}\label{group}
      \displaystyle\min_{\delta} & f(x_0+\delta, t) + \lambda ||\delta||_{2, 0}\\
         s.t. & ||\delta||_{\infty}\le \epsilon, \,\, 0\le x_0+\delta\le 1,
    \end{array}\end{equation}
where $||\delta||_{2, 0}$  is defined as
    $$||\delta||_{2,0}=|\{i: ||\delta_{G_i}||\not= 0, i=1, 2, \cdots, m\}|.$$
By similarly using the indicator function $I_{[l, u]}(\delta)$, problem \eqref{group} can be further written equivalently as
      \begin{equation}\label{gregul}
          \displaystyle\min_{\delta}  f(x_0+\delta, t) + \lambda ||\delta||_{2, 0}+I_{[l, u]}(\delta).
      \end{equation}

Suppose that $f$ is a smooth nonconvex function, whose gradient is Lipschitzian with   Lipschitz constant $L$. Similar to \eqref{equ3}, given $\delta^k$, we consider the problem
      \begin{equation}\label{gequ3}
         \begin{array}{l}
          \displaystyle\min_{\delta} \frac L2 ||\delta-[\delta^k-\frac 1L \nabla_{\delta} f(x_0+\delta^k, t)]||^2 + \lambda ||\delta||_{2, 0}\\
          ~~~~~~~~~ +I_{[l, u]}(\delta)\\
          =\displaystyle\min_{\delta} \frac L2 ||\delta-s_L(\delta^k)||^2 + \lambda ||\delta||_{2, 0}+I_{[l, u]}(\delta).
          \end{array}
      \end{equation}

In \cite{beck19}, the authors presented the solution to a problem more general than  \eqref{gequ3}. Although their derivation can be applied, we still exhibit the concise solution to problem \eqref{gequ3} and give a simple proof, for completeness.

\begin{proposition}
The minimum solution of problem \eqref{gequ3} is
   \begin{equation}\label{IHT}
      \delta_{G_i}^{k+1}=\left\{ \begin{array}{cl}
         [\Pi_{[l, u]}(s_L(\delta^k))]_{G_i}, & \text{~~if~} ||[s_L(\delta^k)]_{G_i}||^2-\\
         & ||[\Pi_{[l, u]}(s_L(\delta^k))-s_L(\\
         & \delta^k)]_{G_i}||^2 > \frac{2\lambda}L;\\
          0, & \text{~~otherwise,}\\
      \end{array}\right.
   \end{equation}
\end{proposition}
for $i=1, 2, \cdots, m$.
\begin{proof}
First, due to group-wise separability, Equation \eqref{gequ3} can be decomposed into $m$ subproblems ($i=1, 2, \cdots, m$):
     \begin{equation}\label{gsubp}
       \displaystyle\min_{\delta_{G_i}} \frac L2 ||[\delta-s_L(\delta^k)]_{G_i}||^2 + \lambda\cdot  \text{sgn}(\|\delta_{G_i}\|)+I_{[l_{G_i}, u_{G_i}]}(\delta_{G_i}),
     \end{equation}
If $||\delta_{G_i}||=0$, which means $\delta_{G_i}=0$, then the objective value of problem \eqref{gsubp} is  $\frac L2 ||[s_L(\delta^k)]_{G_i}||^2$;

If $||\delta_{G_i}||\not=0$, then $\text{sgn}(\|\delta_{G_i}\|)=1$, and problem \eqref{gsubp} becomes
     \begin{equation}\label{subp1}
       \lambda+ \sum_{j\in G_i} \displaystyle\min_{\delta_j} \frac L2 (\delta_j-[s_L(\delta^k)]_j)^2 + I_{[l_j, u_j]}(\delta_j).
     \end{equation}
In \eqref{subp1}, it is obvious that the minimum solution is $[\Pi_{[l, u]}(s_L(\delta^k))]_j, ~~\forall j\in G_i$, which can be unified  as
     $[\Pi_{[l, u]}(s_L(\delta^k))]_{G_i}.$
And the minimum value of \eqref{subp1} is
    \begin{equation*}
       \lambda+ \sum_{j\in G_i} \frac L2 ([\Pi_{[l, u]}(s_L(\delta^k))-s_L(\delta^k)]_j)^2,
     \end{equation*}
which can be written simply  as $\lambda+ \frac L2 ||[\Pi_{[l, u]}(s_L(\delta^k))-s_L(\delta^k)]_{G_i}||^2$.  Thus by comparing the two objectives, the solution of problem \eqref{gequ3} can be obtained as \eqref{IHT}.
\end{proof}

Now, let $g(\delta)=\lambda ||\delta||_{2, 0}+I_{[l, u]}(\delta)$. We denote the solution of problem \eqref{gequ3}  as
    \begin{equation*}\begin{array}{l} 
    \text{Prox}_{\frac 1L g}(\delta^k-\frac 1L \nabla_{\delta} f(x_0+\delta^k, t))\\
    =\arg\!\displaystyle\min_{\delta}  \frac L2 ||\delta-[\delta^k-\frac 1L \nabla_{\delta} f(x_0+\delta^k, t)]||^2 + \lambda ||\delta||_{2, 0}\\
    ~~~~~~~~~~~~~ +I_{[l, u]}(\delta).
    \end{array}\end{equation*}
Note that $g(\delta)=\lambda ||\delta||_{2, 0}+I_{[l, u]}(\delta)$ is also a proper and lower semicontinuous function, the convergence of nmAPG to a critical point can similarly be assured just like in the element-wise sparsity case.

\subsection{Putting Homotopy into Our Solution}


Existing sparse adversarial attack methods that use regularized optimization generally pre-select an ad-hoc fixed weight for the regularization term that controls sparsity \cite{Fan2020ECCV,xu2018structured}. This setting is over-idealistic, and in practice requires multiple trial-and-errors on tuning the weight in order to achieve certain desired sparsity level. To alleviate this issue, we format our nmAPG based algorithm in a \textbf{homotopy} manner \cite{doi:10.1137/120869997,lin2014adaptive}, which allows us to optimize using a sequence of decreasing weights for regularizers. It can keep a multi-stage homotopy solution; in each stage, we optimize \eqref{equ1} using nmAPG until
it reaches the maximum inner iterations. That can continuously provide a sparse initial solution from the last stage for the next stage optimization, leading to the final sparse solution.

Despite homotopy being a highly effective and flexible strategy for convex optimization, it might not be directly applied to highly nonconvex problems with high dimension such as the adversarial attack of a DNN. For example, even if an input perturbation into the nmAPG for problem \eqref{equ1} is sparse, the converged solution by nmAPG might be a bad local minimum and might not be sparse, since the optimized function is highly nonconvex. Below, we describe several strategies that we propose to stabilize and improve the homotopy algorithm for our problem.

\textbf{Initial Weight Search:} The initial value of the weight {$\lambda$} is important for subsequent performance.  For homotopy, an ideal initial weight in problem \eqref{equ1} should produce a solution $\delta$ near 0 in the initial iteration of nmAPG. This can be simply determined by increasing its value in a coarse-grained manner until the  nmAPG with only the first iteration  gives  $\delta=0$, since if $\lambda$ is large enough, nmAPG will focus on optimizing the sparsity function. Then the value is decreased
in a relatively fine-grained manner until the first iteration of nmAPG starts to update on {$\delta$}.  The obtained value is finally multiplied by a constant $c$  and is assigned as the initial weight $\lambda_0$, to better enforce sparsity constraint. Algorithm \ref{lambdaSearch} summarizes our strategy in details.



\begin{algorithm}[h]
\caption{Our Subroutine for Initial Weight Search (Lambda\_Search)}
\label{lambdaSearch}
\begin{algorithmic}[1]{\small
\INPUT $x_0$: benign image; $t$: target; \\
      $eta$, $delta$, $rho$: parameters of nmAPG;\\
     $c$, $v$, $\beta$: parameters of this algorithm.
\OUTPUT $\lambda$.
\STATE {$\delta^0=0$;} {$\lambda=\beta$;}
\REPEAT
\STATE $\delta^1$=nmAPG($\delta^0$, $x_0$, $t$, $\lambda$, $eta$, $delta$, $rho$, $v$, MaxIter=1);

\STATE {$\lambda=\lambda+\beta$;}
\UNTIL{$\delta^1=0$;}

\REPEAT
\STATE $\delta^1$=nmAPG($\delta^0$, $x_0$, $t$, $\lambda$, $eta$, $delta$, $rho$, $v$, MaxIter=1);

\STATE {Decrease $\lambda$ by a factor;}
\UNTIL{$\delta^1 \not= 0$;}
\STATE {$\lambda = \lambda * c$.}}
\end{algorithmic}
\end{algorithm}

\textbf{Additional Sparsity Control by $\ell_0$ Norm Changes:} Even with a good initial weight, the homotopy solution with $\ell_0$ regularizer remains to be sensitive to the regularization parameter $\lambda$ along the homotopy path,
which may not suffice for precise sparsity control for problem \eqref{equ1} when both the loss function and the regularizer are nonconvex.  Moreover, due to the same reason,
for the problem \eqref{equ1} with a fixed weight $\lambda$, even the input solution of nmAPG  is sparse, the converged solution  might not be sparse. We hereby introduce an additional control for sparsity.

Specifically, we constrain the maximum number of $\ell_0$ changes to be $v$  in every outer iteration of homotopy.
Suppose {$r$} is the $\ell_0$-norm of the input perturbation {$\delta$} to  nmAPG in the current outer iteration of homotopy.  Then, at the end of  every inner iteration of nmAPG, we simply keep the entries of {$\delta$} with the top $(r + v)$ absolute values and reduce all other entries to $0$. This will keep the number of nonzero entries of the perturbation  increased steadily, which avoids the situation that a small decrease of the weight $\lambda$ makes the $\ell_0$-norm of the perturbation increase suddenly.



%
%


The value of $v$ is an input constant of our homotopy algorithm. However, it will be updated temporarily as a small positive integer to be input to nmAPG,  whenever the current perturbation $\delta^{k+1}$ satisfies that
  \begin{equation}\label{ineq}
     \frac{||\delta^{k+1}||_1} {||\delta^{k+1}||_0} \le \epsilon * \gamma,
  \end{equation}
where $\gamma<1$ is a user-specified constant. An intuition behind this mechanism is that, if inequality \eqref{ineq} is satisfied, then the average perturbation magnitude of $\delta^{k+1}$ is much smaller than the invisible threshold $\epsilon$, which means the number of $||\delta^{k+1}||_0$ entries has not been fully used to attack the classification model at present. Hence $v$ is set temporarily to a small positive integer, and nmAPG will solve problem \eqref{equ1} with a smaller $v$ until inequality \eqref{ineq} is violated.


%


\textbf{Optional Post-Attack Perturbation:} The above scheme can already generate sparse adversarial perturbations. However, we observe that our homotopy algorithm sometimes gets stuck in bad local minima, which is inevitable due to the nonconvex $f$.  In this case, as $\lambda$ decreases in the homotopy algorithm,
the $\ell_1$-norm of the perturbation increases disproportionally to the corresponding $\ell_0$-norm, which means the algorithm cannot effectively use the perturbed entries, and inequality \eqref{ineq} is satisfied.
So inequality \eqref{ineq} is regarded as a trigger condition that our algorithm falls into a local minimum after the nmAPG stage.

To alleviate this issue, we intend to provide some ``push" to help our algorithm escape from the local minimum faster. In each outer iteration of the homotopy,  we check whether  inequality \eqref{ineq} is satisfied.
If it is the case, we propose to execute the following post-attack perturbation step:
    \begin{equation}\begin{array}{ll}\label{equ4}
      \displaystyle\min_{\delta} & w_1 f(x_0+\delta, t) + w_2
      ||\delta||_p\\
         s.t. & ||\delta||_{\infty}\le \epsilon, \,\,0\le x_0+\delta\le 1,
    \end{array}\end{equation}
where  $w_1 \gg w_2$, $p$ can be either 1, 2, or $\infty$, and {$\delta$} is now enforced only on \textbf{nonzero entries} of the output perturbation of the nmAPG stage.

The assumption that $w_1 \gg w_2$ makes an optimizer focus on minimizing the function $f$, and the value of $||\delta||_p$ might increase.  Then the inequality \eqref{ineq} might be violated, which means our algorithm might have escaped from the bad local minimum.
Since we only need to provide some push directions rather than precise updates, we simply optimize problem \eqref{equ4} using gradient descent for a  number of iterations proportional to the current $\ell_0$. After the post attack stage, combined with the nmAPG stage with smaller $v$ in the next outer iteration of homotopy, the $\ell_1$-norm of the perturbation would increase such that inequality \eqref{ineq} is violated, and then we can set $v$ back to normal.


Our full \textbf{homotopy attack} algorithm are summarized as Algorithm \ref{attack}. Starting with the initial weight search, the algorithm has an outer iteration by homotopy. Within each outer iteration, the main step is to optimize an $\ell_0$-regularized loss by leveraging nmAPG (which will have its inner iterations), followed by an additional $\ell_0$ change control step, and an optional post-attack to escape from bad local minima.

\begin{algorithm}[htpb]
\caption{The Homotopy Attack Algorithm}
\label{attack}
\begin{algorithmic}[1]{\small
\INPUT $x_0$: benign image; $t$: target; $\epsilon$: invisible threshold; \\
       $eta$, $delta$, $rho$, MaxIter: parameters of nmAPG;\\
      $c$, $v$, $\beta$, $\gamma$: parameters of this algorithm.
\OUTPUT $\delta$.
\STATE {$\delta^0=0$;} {$v_{ini}=v$;} {$k=0$;}
\STATE {$\lambda=$Lambda\_Search($x_0$, $t$, $c$, $eta$, $delta$, $rho$, $v$, $\beta$);}

\REPEAT
\STATE {$\delta^{k+1} = $ nmAPG($\delta^k$, $x_0$, $t$, $\lambda$, $eta$, $delta$, $rho$, $v$, MaxIter);}
\STATE{$v=v_{ini}$;}
\IF{not success}
\IF{$||\delta^{k+1}||_1 \le ||\delta^{k+1}||_0 * \epsilon * \gamma$}
\STATE{Set $v$ as a small integer;}
\STATE{Conduct post-attack perturbation;}
\ENDIF
\STATE {Decrease $\lambda$ by a factor;}
\ENDIF
\STATE{$k = k + 1$;}
\UNTIL Attack Succeeds.}
\end{algorithmic}
\end{algorithm}

\vspace{-5 mm}



\section{Experiments}
In this section, we conduct comprehensive experiments with diverse setups to validate the effectiveness of proposed homotopy algorithm on the CIFAR-10 \cite{Krizhevsky2009LearningML} and the ImageNet \cite{5206848} datasets. Section~\ref{sec:setup} collects our experimental setups. Then, we compare multiple previous state-of-the-art (SOTA) sparse adversarial attack approaches in Section~\ref{sec:tar_adv}, ~\ref{sec:untar_adv}, and ~\ref{subsec_add}, including GreedyFool \cite{dong2020greedyfool}, SAPF \cite{Fan2020ECCV}, and StrAttack \cite{xu2018structured}. Moreover, ablation studies, the application to group-wise sparsity, empirical time cost, and visualization are provided in Section~\ref{sec:ablation}, ~\ref{subsec_gs}, ~\ref{subsec_time}, and ~\ref{subsec_vis}, respectively.

\subsection{Experiment Setup}\label{sec:setup}

\textbf{Dataset and Classification Model Setting:}
Following the practice of \cite{xu2018structured} and \cite{Fan2020ECCV}, we randomly select 1000 images from the test set of CIFAR-10  for targeted attack. Each selected image would  be attacked with 9 target classes except its own class label, thus 9000 adversarial examples are generated in total. As for the ImageNet dataset, we select randomly 100 images from the validation set and  9 classes as the targets for targeted attack, so we generate 900 adversarial examples from the ImageNet dataset. For nontargeted attack, we randomly select 5000 images from the test set of CIFAR-10, and 1000 images from the validation set of ImageNet as the input images.

For the classification model on CIFAR-10 dataset, we follow the practice of \cite{Fan2020ECCV}, \cite{DBLP:journals/corr/CarliniW16a}, and \cite{xu2018structured} to train a network that consists of four convolution layers, two max-pooling layers, and two fully-connected layers. Our trained network can achieve 80\% classification accuracy on the CIFAR-10 dataset. For the classification model on the ImageNet dataset, we choose to use the pretrained Inception-v3 model \cite{Incep2016}, which can achieve 77.45\% top-1 classification accuracy and 96\% top-5 classification accuracy on the ImageNet dataset.

\textbf{Parameter Setting: }
For targeted attack, the adversarial loss function $f()$ is set as the cross entropy function. For nontargeted attack, we adopt the loss function proposed in \cite{DBLP:journals/corr/CarliniW16a} directly: 
\begin{equation*}
f(x_0, y_0)=\max \{D(x_0)_{y_0} - \max_{i\not= y_0} \{ D(x_0)_i\},  -\kappa\},
\end{equation*}
where {$x_0$} is a benign image and {$y_0$} is its ground-truth class label, $D$ is the model to attack, and ${D}(x_0)_{y_0}$ is the logit for class $y_0$. The confidence parameter $\kappa$ is set to $0$.

Since we are highly interested in generating sparse and invisible adversarial perturbations while not extremely sparse but visible ones, 
we maintain a relatively small $\ell_\infty$-norm of  generated perturbations. That is, we set $\epsilon$ to  $0.05$, which is a relatively small number in the $[0, 1]$ range of a valid image.

While, except GreedyFool \cite{dong2020greedyfool}, the other two compared methods \cite{Fan2020ECCV, xu2018structured} do not have a control on the $\ell_\infty$ of  generated perturbations. They either use a fixed number as an upper bound on the $\ell_0$-norm, or adopt a fixed weight for sparsity regularizer. When enforcing a strict sparsity constraint, their methods may fail or cause the $\ell_\infty$ of generated perturbations to be extremely large, which is perceptible by human eyes. Therefore, to establish a fair comparison, we use our
$\epsilon$ setting as a reference, and tune the parameters of their methods to ensure that the average $\ell_\infty$ of generated perturbations is close to  $\epsilon$. For GreedyFool \cite{dong2020greedyfool}, we use their original implementation and set the upper bound on $\ell_\infty$ to 0.05. The similar level of $\ell_\infty$ enables us to make a fair comparison on the $\ell_0$ of generated perturbations.

Other parameters of our algorithm and the parameters for group-wise sparsity are discussed in detail in the appendix.

\begin{table*}[t] 
  \centering
  \caption{Statistics of attack success rate and  average $\ell_p$-norms ($p = 0,1,2, \infty$) of targeted attack on  CIFAR-10 and ImageNet datasets.}
  \label{table:tar_result}
  \scalebox{0.8}[0.8]{
\setlength{\tabcolsep}{0.9 mm}{
  \begin{tabular}{|l|c|ccccc|ccccc|ccccc|}
  \hline
       \multirow{2}{*}{Database} & \multirow{2}{*}{Method} &
       \multicolumn{5}{c}{Best case}  &
       \multicolumn{5}{c}{Average case}  &
       \multicolumn{5}{c|}{Worst case}  \\
  \cline{3-17}
 & & ASR & $\ell_0$ & $\ell_1$ & $\ell_2$ & $\ell_{\infty}$	 & ASR & $\ell_0$ & $\ell_1$ & $\ell_2$ & $\ell_{\infty}$ & ASR & $\ell_0$ & $\ell_1$ & $\ell_2$ & $\ell_{\infty}$\\
\hline
\hline
\multirow{4}{*}{CIFAR-10}

      & GreedyFool    & 100&199  &6.583&0.468 &0.048 &100 &289 & 10.037 &0.606 & 0.049 & 100 &445 &15.935 &0.786 &0.049 \\
      \cline{2-17}
      & SAPF    & 100&314 & 4.067& 0.258& 0.051& 100&518 &7.487 &0.362 &0.051 &100 &594 &9.860 & 0.452&0.051 \\
      \cline{2-17}
      & StrAttack    &100 & 363&4.570 &0.248 & 0.050& 100& 546&8.267 & 0.501&0.054 &100 & 657&10.843 & 0.485& 0.056\\
      \cline{2-17}
      & Homotopy-Attack     & 100&110 & 5.166&0.467 &0.047 &100 &165 &7.955 &0.580 &0.049 & 100& 234& 11.310&0.686 &0.049 \\
\hline
\hline
\multirow{4}{*}{ImageNet}

      & GreedyFool    & 100&8937 &283.705 & 3.287&0.049 &100 &10837 &290.848 &3.176 &0.049 &100 &12197 &330.049 &3.552 &0.049 \\
      \cline{2-17}
      & SAPF   &100 & 37923& 84.726&0.852 &0.056 &100 &54743 &117.488 &0.980 &0.054 &100 &60374 &143.718 &1.179 & 0.062\\
      \cline{2-17}
      & StrAttack    &100 & 39593& 95.122& 1.039& 0.061& 100& 55410&136.314 &1.144&0.056  &100 &61328 &165.434 &1.203 & 0.067\\
      \cline{2-17}
      & Homotopy-Attack     &100 & 2399& 99.178&2.030 &0.049 & 100&2706 & 111.697&2.134 &0.049 &100 &3065 & 126.933&2.288 &0.050 \\
  \hline
  \end{tabular}
 }
  }
\end{table*}

\textbf{Evaluation Metrics: }
We report the average $\ell_p$-norms of  generated perturbations, where $p = 0, 1, 2, \infty$, and the attack success rates (ASR) of the experimented methods with both the targeted and nontargeted settings. In the targeted setting, we follow 
\citet{Fan2020ECCV} and \citet{xu2018structured} to separate the results into three cases: the best case, the average case and the worst case. The best case presents the results of target class that leads to the most sparsity, the worst case shows the results of target class that leads to the least sparsity, and the average case gives the average results of all 9 target classes.

\subsection{Targeted Attack}\label{sec:tar_adv}
\label{tar_sec}
Table \ref{table:tar_result} presents the results of targeted attack on the CIFAR-10 
and ImageNet, where ``Homotopy-Attack" denotes our proposed algorithm.
From this table, we can see that all compared methods can  achieve 100\% attack success rate. However, our algorithm outperforms the other compared methods on sparsity level by a large margin. Specifically, our algorithm can generally achieve 5.4\% sparsity on {$32 \times 32 \times 3$} CIFAR-10 images and 1\% sparsity on {$299 \times 299 \times 3$} ImageNet images in average case, which is   42.91\% fewer  on CIFAR-10 and 75.03\% fewer on ImageNet, respectively, when compared to SOTA.



From Table \ref{table:tar_result}, a noticeable finding is that, the ratios of $\ell_1$ to $\ell_0$ of our algorithm are generally larger than the other three methods on both the two datasets. We believe this is not a bad phenomenon. To enforce higher level of sparsity, there needs to be trade-offs between the $\ell_0$-norm and the $\ell_1$ and $\ell_2$-norms. Further, we can discover from Table \ref{table:tar_result} that,  the average ratios of $\ell_1$ to $\ell_0$  of our generated perturbations are generally higher than 0.04, which is very close to the  $\ell_\infty$-threshold $\epsilon$. This result indicates that  our algorithm is able to make the most trade-offs between the $\ell_0$-norm and the $\ell_1$ and $\ell_2$-norms, which can lead to sparser results. This may also indicate that relaxing the $\ell_0$ constraint to $\ell_1$ may not be optimal in the case of sparse adversarial attack.

\subsection{Nontargeted Attack}\label{sec:untar_adv}

GreedyFool \cite{dong2020greedyfool}  outperforms the state-of-the-art methods like SparseFool \cite{Modas_2019_CVPR} and {PGD$_0$} \cite{Croce_2019_ICCV} by a significant large margin on nontargeted attack.  Since SAPF \cite{Fan2020ECCV} adopted a similar optimization method and achieved better results than StrAttack \cite{xu2018structured} in Table \ref{table:tar_result}, we focus on comparing compare our algorithm with GreedyFool and SAPF on nontargeted attack.

Experimental results of nontargeted attack on the ImageNet 
and CIFAR-10 
datasets are listed in Table \ref{table:nontar_result}. From this table, we can see that all methods achieve 100\% attack success rate. Looking at the $\ell_p$-norms, we can find that our algorithm on average only need to perturb 2.3\% entries on the {$32 \times 32 \times 3$} images from CIFAR-10 dataset, and   0.14\% entries on the {$299 \times 299 \times 3$} images from ImageNet dataset, which is 37.93\% fewer on CIFAR-10 and 65.69\% fewer on ImageNet, respectively, when compared to SOTA.

Further, we can find from Table \ref{table:nontar_result} that, the ratios of $\ell_1$ to $\ell_0$ of our algorithm are still larger than 0.04, and our algorithm can generate much sparser result. Thus the findings  mentioned in Subsection \ref{tar_sec} also apply for nontargeted attack.

\begin{table}[h]  
  \centering
  \caption{Statistics of attack success rate and average $\ell_p$-norms ($p = 0,1,2, \infty$) of nontargeted attack on CIFAR-10 and ImageNet.}
  \label{table:nontar_result}
  \scalebox{0.7}[0.7]{
\setlength{\tabcolsep}{2.5 mm}{
  \begin{tabular}{|l|c|ccccc|}
  \hline
   Database & Method &  ASR & $\ell_0$ & $\ell_1$ & $\ell_2$ & $\ell_{\infty}$ \\
 \hline
\hline
\multirow{3}{*}{CIFAR-10}
      & GreedyFool  & 100 & 116 & 4.395 & 0.392 & 0.048  \\
  \cline{2-7}
      & SAPF     & 100 &   352 &  6.070 &  0.365 &   0.056 \\    
      \cline{2-7}
      & Homotopy-Attack     & 100 & 72 & 3.427 &0.367 &0.049  \\
\hline
\hline
\multirow{3}{*}{ImageNet}

      & GreedyFool      & 100 &1128  & 35.764 &0.964  & 0.049 \\
      \cline{2-7}
      & SAPF     & 100 &  2063 &  24.907 &   0.637 &   0.058 \\   
      \cline{2-7}
      & Homotopy-Attack       & 100  &387  & 16.387 &0.685 &0.047   \\

  \hline
  \end{tabular}
 }
  }
\end{table}

\subsection{Scale Up to Stronger Backbone Network}\label{subsec_add}
To further validate our method's effectiveness on CIFAR-10 dataset, we train a ResNet-18 \cite{7780459} which can achieve 93\% accuracy on CIFAR-10 dataset, and conduct the previous experiments with GreedyFool \cite{dong2020greedyfool}  and SAPF \cite{Fan2020ECCV}. The results are listed in Tables \ref{table:resnet}. From the table, we can see that our algorithm outperforms GreedyFool (SOTA) by a larger margin.  Our algorithm is able to reduce 61.46\% sparsity, and  62.87\% sparsity respectively on nontargeted and targeted attacks when compared to GreedyFool. It indicates that our algorithm is more effective than greedy-based methods when attacked networks become stronger.

\begin{table}[h!]   
  \centering
  \caption{Statistics of targeted and nontargeted attacks on ResNet-18 on CIFAR-10.}
  \label{table:resnet}
  \scalebox{0.7}[0.7]{
\setlength{\tabcolsep}{2.5 mm}{
  \begin{tabular}{|l|c|ccccc|}
  \hline
   Attack Type & Method &  ASR & $\ell_0$ & $\ell_1$ & $\ell_2$ & $\ell_{\infty}$ \\
 \hline
\hline
\multirow{3}{*}{Nontargeted}
      & GreedyFool  & 100 & 192 & 6.192 & 0.439 & 0.048  \\
      \cline{2-7}
      & SAPF     & 100 &   723 &  7.758 &   0.345 &   0.055 \\      
      \cline{2-7}
      & Homotopy-Attack & 100 & 74 & 3.405 &0.365 &0.047  \\
        \hline
\multirow{3}{*}{Targeted}
      & GreedyFool  & 100 & 668 & 19.667 & 0.773 & 0.048  \\
      \cline{2-7}
      & SAPF     & 100 &   1320 &   11.801 &   0.437 &   0.071 \\      
      \cline{2-7}
      & Homotopy-Attack  & 100 & 248 & 11.359 &0.676 &0.049  \\
  \hline
  \end{tabular}
 }
  }  
\end{table}


\subsection{Ablation Study}\label{sec:ablation}


To further understand the contribution of each component in our algorithm, ablation studies are conducted on the CIFAR-10 and ImageNet datasets. Without loss of generality, we only experiment on targeted attack, since it is more difficult than nontargeted attack. The results are listed in Table  \ref{table:component_result}, in which we term the homotopy algorithm without additional sparsity control and optional post attack stage as ``Pure-Homotopy", and the nmAPG algorithm with binary weight search as ``nmAPG".


From Table \ref{table:component_result}, we can see  that both nmAPG with binary weight search and homotopy without additional sparsity control or post attack stage perform much worse than our homotopy attack algorithm. The sparsity levels are improved  significantly after  adding the additional control and post attack stage to Pure-Homotopy on both datasets. We believe there are 2 reasons behind this: (1) The $\ell_0$-norm regularizer is nonsmooth and nonconvex, and it is sensitive to the change of weight {$\lambda$}  in either the homotopy manner or the binary search manner. So we cannot be certain about if the decrease of $\lambda$ in each iteration of the homotopy algorithm will cause  a significant change of the $\ell_0$-norm, which we do not want it to happen. So further control of the maximum increase of $\ell_0$ is needed here. (2) Since the loss function $f$ is nonconvex, the optimization can easily be trapped in local minima. If that happens, we need to either push it out or give it some directions. Without the ``push", the algorithm would require a huge increase of the $\ell_0$-norm to escape. So the post attack stage is needed.

\begin{table}[h] \vspace{-2mm}
  \centering
  \caption{Statistics of  attack success rate and  average $\ell_p$-norms ($p = 0,1,2, \infty$) of component analysis experiment.}
  \label{table:component_result}
  \scalebox{0.68}[0.68]{
\setlength{\tabcolsep}{2.5 mm}{

  \begin{tabular}{|l|c|ccccc|}
  \hline
   Database & Method &  ASR & $\ell_0$ & $\ell_1$ & $\ell_2$ & $\ell_{\infty}$ \\
 \hline
\hline
\multirow{2}{*}{CIFAR-10}
      & Pure-Homotopy  & 100 & 223 & 10.087 & 0.644 & 0.050  \\
      \cline{2-7}
      & nmAPG     & 100 &   283 &   12.687 &   0.714 &   0.050 \\
      \cline{2-7}
      & Homotopy-Attack     &100 &168 &7.409 &0.555 &0.049  \\
\hline
\hline
\multirow{2}{*}{ImageNet}

      &   Pure-Homotopy     & 100& 16343& 340.691&3.094 &0.050  \\
      \cline{2-7}
      & nmAPG     & 100 &   13582 &  303.195 &  2.998 &   0.050 \\
      \cline{2-7}
      & Homotopy-Attack       & 100& 2889& 117.97&2.209 &0.049  \\
  \hline
  \end{tabular}
 }
  }
\vspace{-2mm}
\end{table}

\subsection{Group-wise Sparsity}\label{subsec_gs}

In this subsection, we show the group-wise sparsity experimental results of our algorithm on targeted attack. The results are given in Table \ref{table:group_result}. Since we only constrain group-wise sparsity, the average $\ell_0$-norm is larger than the corresponding one in Table \ref{table:tar_result}, where we conduct pixel-wise sparsity. However, we can see from the $\ell_{2,0} $ that, the group-wise sparsity level is very high. Specifically, we can on average achieve 6.88\% and 2.05\% group sparsity on CIFAR-10 and ImageNet, respectively.

\begin{table}[h] 
  \centering
  \caption{Statistics of our algorithm on group-wise sparse targeted attack.}
  \label{table:group_result}
  \scalebox{0.66}[0.66]{
\setlength{\tabcolsep}{2.5 mm}{
  \begin{tabular}{|l|ccccccc|}
  \hline
   Database &  ASR & $\ell_0$&$\ell_{2,0} $ & $\ell_1$ & $\ell_2$ & $\ell_{\infty}$ & no. of groups \\
 \hline
\hline
\multirow{1}{*}{CIFAR-10}
       &100  &502 & 5.501&19.407 &0.881 &0.047 & 80\\
\hline
\multirow{1}{*}{ImageNet}
        & 100 &7711 & 8.667&292.862 & 3.461&0.048 & 423\\
  \hline
  \end{tabular}
 }}
\end{table}

\subsection{Empirical Time Cost}\label{subsec_time}

In this subsection, we compare the running time of our algorithm with state-of-the-art methods. The results are given in Table \ref{table:time_result}. From the table we can see that our algorithm is much faster than SAPF \cite{Fan2020ECCV} and StrAttack \cite{xu2018structured}, since our algorithm does not require multiple reruns of the whole algorithm with binary searched weights to ensure attack success and result quality. 

GreedyFool \cite{dong2020greedyfool}, on the other hand, runs faster than our algorithm because of  its greedy nature. However, previous experiments in Tables \ref{table:tar_result}, \ref{table:nontar_result}, and \ref{table:resnet} show that our algorithm always generate much sparser perturbations with lower $\ell_1$ and $\ell_2$ norms than Greedyfool under the same $\ell_\infty$ constraints.

\begin{table}[h]  
  \centering
  \caption{Empirical time cost of targeted attack in seconds.}
  \label{table:time_result}
  \scalebox{0.66}[0.66]{
\setlength{\tabcolsep}{2.5 mm}{
  \begin{tabular}{|l|cccc|}
  \hline
   Database &  GreedyFool & Homotopy Attack & SAPF & StrAttack  \\
 \hline
\hline
\multirow{1}{*}{CIFAR-10}
       & 0.43 & 22.64 &  100.15 &  142.65 \\
\hline
\multirow{1}{*}{ImageNet}
        &  71.49 &  516.57 &  1589.54 &  2242.78 \\
  \hline
  \end{tabular}
 }}
\end{table}

\subsection{Visualization}\label{subsec_vis}
\begin{figure}[h]
  \vspace{-2mm}
\centering
    \subfigure 
    {
        \includegraphics[height=1.8cm]{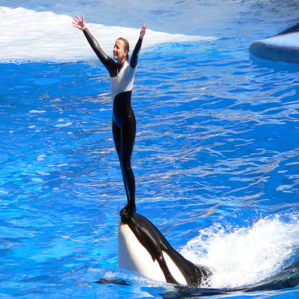}
    }
    \subfigure 
    {
        \includegraphics[height=1.8cm]{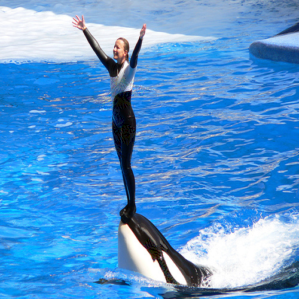}
    }
    \subfigure 
    {
        \includegraphics[height=1.8cm]{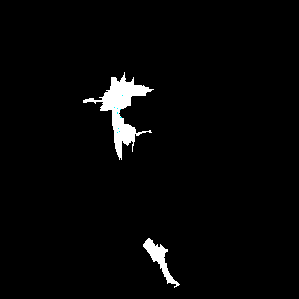}
    }
    \subfigure 
    {
        \includegraphics[height=1.8cm]{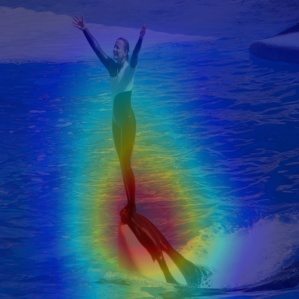}
    }
\vskip -0.3cm
     \subfigure 
     {
        \includegraphics[height=1.8cm]{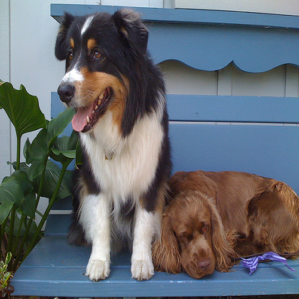}
    }
        \subfigure 
    {
        \includegraphics[height=1.8cm]{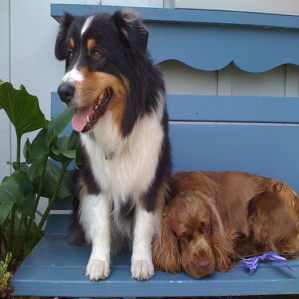}
    }
    \subfigure 
    {
        \includegraphics[height=1.8cm]{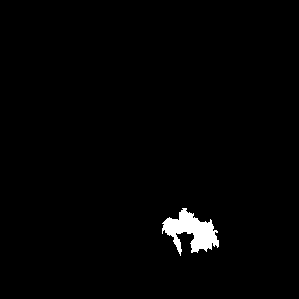}
    }
    \subfigure 
    {
        \includegraphics[height=1.8cm]{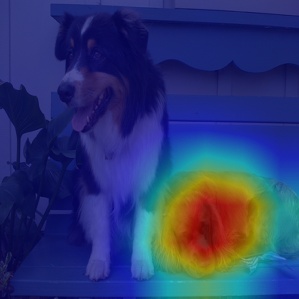}
    }

\vskip -0.3cm
     \subfigure 
     {
        \includegraphics[height=1.8cm]{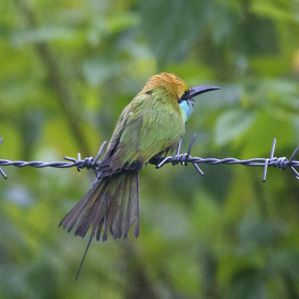}
    }
        \subfigure 
    {
        \includegraphics[height=1.8cm]{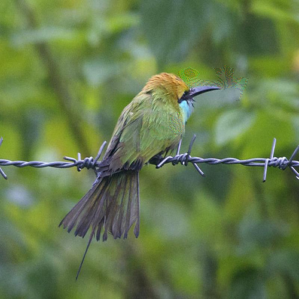}
    }
    \subfigure 
    {
        \includegraphics[height=1.8cm]{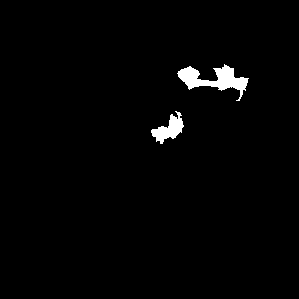}
    }
    \subfigure 
    {
        \includegraphics[height=1.8cm]{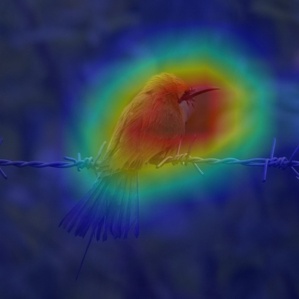}
    }
\vskip -0.3cm
     \subfigure 
     {
        \includegraphics[height=1.8cm]{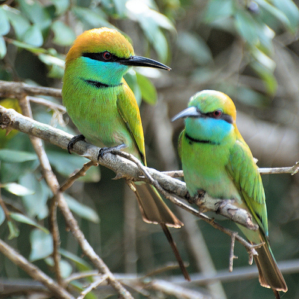}
    }
        \subfigure 
    {
        \includegraphics[height=1.8cm]{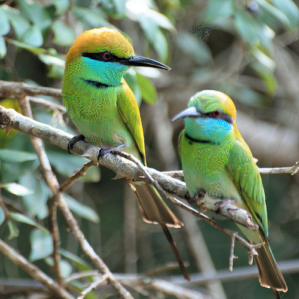}
    }
    \subfigure 
    {
        \includegraphics[height=1.8cm]{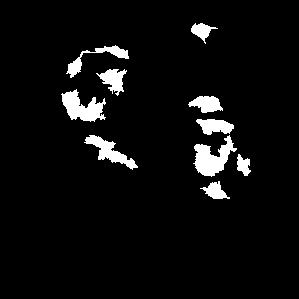}
    }
    \subfigure 
    {
        \includegraphics[height=1.8cm]{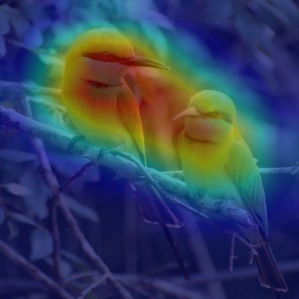}
    }
\vskip -0.3cm
     \subfigure 
     {
        \includegraphics[height=1.8cm]{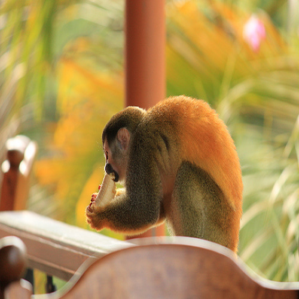}
    }
        \subfigure 
    {
        \includegraphics[height=1.8cm]{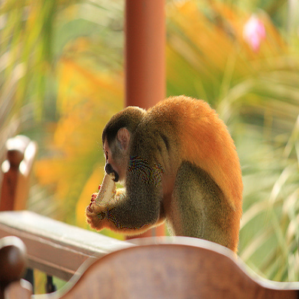}
    }
    \subfigure 
    {
        \includegraphics[height=1.8cm]{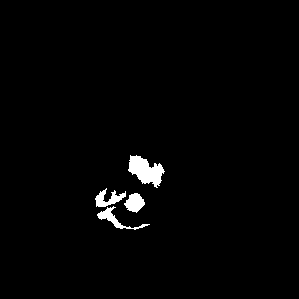}
    }
    \subfigure 
    {
        \includegraphics[height=1.8cm]{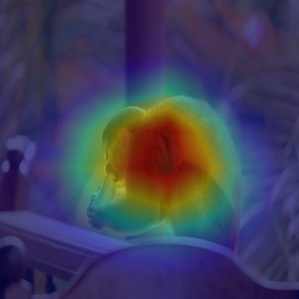}
    }
\vskip -0.3cm
     \subfigure 
     {
        \includegraphics[height=1.8cm]{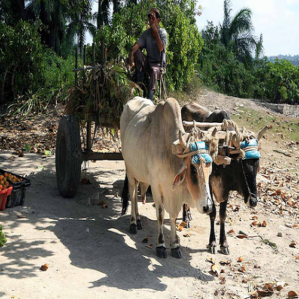}
    }
        \subfigure 
    {
        \includegraphics[height=1.8cm]{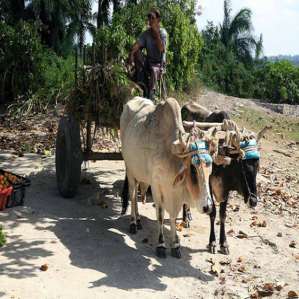}
    }
    \subfigure 
    {
        \includegraphics[height=1.8cm]{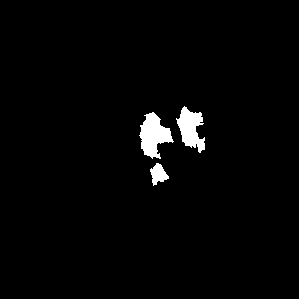}
    }
    \subfigure 
    {
        \includegraphics[height=1.8cm]{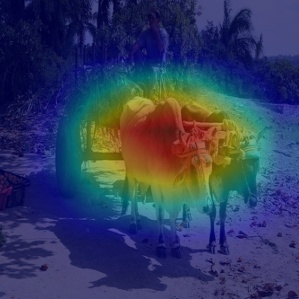}
    }
  \vskip -0.2cm \caption{\small Visualization of targeted attack with group-wise sparsity. Each row from the left to right represents benign image, adversarial example, perturbation positions, and CAM of the original label.  The ground-truth label for each row are killer whale, Sussex spaniel, bee eater, bee eater, titi monkey and oxcart, respectively. The target class is face powder for row 1 and 6, cock for row 2 and 3, and wild boar for row 4 and 5.
  }\label{fig:group}
  \vspace{-5mm}
\end{figure}

The visualizations for pixel-wise sparsity and group-wise sparsity are shown in Figure~\ref{fig:pix}, and~\ref{fig:group} respectively. In Figure \ref{fig:group}, we  use class activation map (CAM) \cite{Zhou_2016_CVPR} to show whether the perturbations generated by our algorithm have a good correspondence with localized class-specific discriminative image regions. In the figures of perturbation positions, black pixels denote no perturbation; white pixels denote perturbing all three RGB channels; pure red, green, blue pixels represent perturbing single channel; and pixels with other color denote perturbing two channels.

From rows  2-6 of Figure~\ref{fig:group}, it is obvious that the generated perturbations cover the most discriminative areas of the objects, thanks to our algorithm achieving high group-sparsity. Row 1, on the other hand, might provide an interesting insight of the mechanism of DNN. It might indicate a high correlation between two different objects (trainer and killer whale) in the training data of DNN. The corresponding CAM also validates this finding.

The reason of perturbations generated with pixel-wise sparsity constraint not always being on discriminative areas like row 3 of Figure~\ref{fig:pix} is that, our algorithm is able to find some pixels not on the object that offer more adversary when enforcing pixel-wise constraint. However, when enforcing group-wise constraint, our algorithm will perturb groups with highest sum adversaries, which generally cover the discriminative areas.

\vspace{5mm}

\section{Conclusion}
In this paper, we have proposed a novel homotopy algorithm  for sparse adversarial attack based on Nonmonotone  Accelerated  Proximal  Gradient Methods for Nonconvex Programming,  an additional control of maximum $\ell_0$ updates and an optional post attack stage per iteration. Extensive experiments show that our algorithm can generate very sparse adversarial perturbations while maintaining relatively low perturbation magnitudes, compared to the state-of-the-art methods. Also, our proposed control of maximum $\ell_0$  updates and the optional post attack stage greatly improve the sparsity level of the homotopy algorithm. 


%

\bibliography{example_paper}
\bibliographystyle{icml2021}

\end{document}